\documentclass[letterpaper, 10 pt, conference]{ieeeconf}  

\IEEEoverridecommandlockouts                              

\usepackage{anyfontsize}
\usepackage{multirow}
\usepackage{graphicx} 
\usepackage{comment}
\usepackage{color}
\usepackage{algorithm}
\usepackage[noend]{algpseudocode} 
\usepackage{amssymb,amsmath}
\newtheorem{theorem}{Theorem}
\newtheorem{lemma}{Lemma}
\newtheorem{definition}{Definition}
\newtheorem{corollary}{Corollary}
\usepackage{hyperref}

\usepackage{ifthen}
\newboolean{shortver}
 \setboolean{shortver}{false}

\title{\LARGE \bf
	Loosely Synchronized Search for Multi-agent Path Finding with Asynchronous Actions
}

\author{Zhongqiang Ren$^{1}$, Sivakumar Rathinam$^{2}$ and Howie Choset$^{1}$
	\thanks{$^{1}$ Zhongqiang Ren and Howie Choset are with Carnegie Mellon University, 5000 Forbes Ave., Pittsburgh, PA 15213, USA. 
	}%
	\thanks{$^{2}$Sivakumar Rathinam is with Texas A\&M University,
		College Station, TX 77843-3123.
	}
}
\begin{document}

\maketitle

\thispagestyle{empty}
\pagestyle{empty}


\begin{abstract}
	Multi-agent path finding (MAPF) determines an ensemble of collision-free paths for multiple agents between their respective start and goal locations.
Among the available MAPF planners for workspace modeled as a graph, A*-based approaches have been widely investigated due to their guarantees on completeness and solution optimality, and have demonstrated their efficiency in many scenarios.
However, almost all of these A*-based methods assume that each agent executes an action concurrently in that all agents start and stop together.
This article presents a natural generalization of MAPF with asynchronous actions (MAPF-AA) where agents do not necessarily start and stop concurrently.
The main contribution of the work is a proposed approach called Loosely Synchronized Search (LSS) that extends A*-based MAPF planners to handle asynchronous actions.
We show LSS is complete and finds an optimal solution if one exists.
We also combine LSS with other existing MAPF methods that aims to trade-off optimality for computational efficiency.
Numerical results are presented to corroborate the performance of LSS and the applicability of the proposed method is verified in the Robotarium, a remotely accessible swarm robotics research platform.
\end{abstract}

\section{Introduction}\label{sec:intro}

\begin{figure}[htbp]
	\centering
	\vspace{-1mm}
	\includegraphics[width=0.9\linewidth]{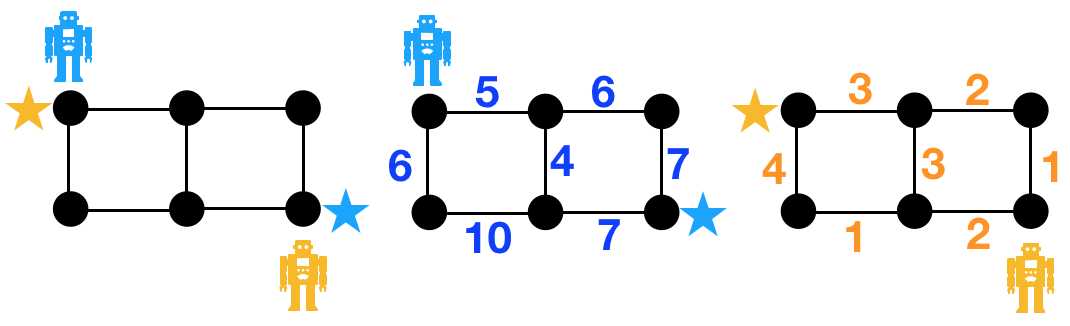}
	\vspace{-4mm}
	\caption{A toy example that illustrates the problem. Blue and yellow numbers indicate the durations required for blue and yellow agents to move through the edges respectively. The goal locations are marked with stars.}
	\vspace{-6mm}
	\label{fig:toy_eg}
\end{figure}

Multi-agent path finding (MAPF), as its name suggests, computes a set of collision-free paths for multiple agents from their respective starts to goal locations. 
Most MAPF methods~\cite{stern2019multi} describe the workspace as a graph, where vertices represent possible locations of agents and edges are actions that move agents between locations.
Conventional MAPF planners \cite{felner2017search,stern2019multi}, including our own~\cite{wagner2015subdimensional}, typically consider the case where each agent executes an action concurrently in that all agents start and stop together.
The requirement of such synchronized actions among agents limits the application of MAPF planners to scenarios where agents move with different speeds.
This paper considers a natural generalization of the MAPF with the agents' \emph{actions running asynchronously,} meaning they do not necessarily start and stop concurrently. We refer to this generalization as MAPF with asynchronous actions (MAPF-AA). In MAPF-AA, different actions by agents may require different time durations to complete. See Fig.~\ref{fig:toy_eg} for a toy example.

Among MAPF planners, A*-based ones, such as HCA*~\cite{silver2005hca}, EPEA*~\cite{goldenberg2014enhanced}, M*~\cite{wagner2015subdimensional}, have been extensively investigated.
These planners provide guarantees on solution completeness and optimality, and outperforms other types of MAPF planners in certain scenarios~\cite{felner2017search}.
However, existing A*-based methods rely on the assumption of synchronous actions to easily identify ``planning steps'' for all agents and run A*-like search.
For MAPF-AA problem, a naive application of conventional A*-based planners may require too fine a discretization of the time dimension so that a common unit time can be found for planners to identify planning steps.
This work aims to overcome the challenge.

The main contribution of this work is a proposed method called Loosely Synchronized Search (LSS) that extends A*-based MAPF planners to handle asynchronous actions. 
In this approach, we introduce a new state space which combines the spatial and temporal information of agents for the purpose of describing asynchronous actions.
The standard state expansion used in A*-based planners is then generalized to also account for the temporal information. Finally, we use dominance principles from the multi-objective optimization literature~\cite{przybylski2017multi,ehrgott2005multicriteria} to compare and prune states that cannot lead to an optimal solution. 
We also prove that LSS is complete and finds an optimal solution if one exists (Sec.~\ref{sec:analysis}).

To show the generality of LSS, we fuse it with M* and recursive M*, which results in LS-M*, LS-rM* (Sec.~\ref{sec:combine_m}). 
We also fuse LS-rM* with Meta-agent Conflict-based Search (MA-CBS)~\cite{sharon2012meta}, which is an algorithm that combines both Conflict-based Search (CBS)~\cite{sharon2015conflict} and A*-based planners to achieve better performance.
We test the algorithms in maps from~\cite{stern2019multi} and our results (Sec.~\ref{sec:result}) show: (1) LS-A* expands far fewer states than a naive adoption of A* for MAPF-AA; (2) LSS can be fused with M* and rM*, resulting in LS-M* and LS-rM* and inflated heuristics improve the computational efficiency while providing bounded sub-optimal solutions; (3) The extension of MA-CBS by leveraging LS-rM* improves the success rates and run time on average when comparing it with the existing CBS-based algorithm~\cite{andreychuk2019multi}, which is, to our limited knowledge, the state-of-the-art search-based planner that can solve MAPF-AA.
Finally, to verify the applicability of our approach to real multi-robot systems, we also execute the paths computed by our planner (LS-rM*) in the Robotarium~\cite{wilson2020robotarium}, a remotely accessible swarm robotics research platform.


\section{Prior Work}\label{sec:prior}
%


MAPF algorithms tend to fall on a spectrum from decentralized to centralized, trading off completeness and optimality for scalability. Finding an optimal solution for MAPF is NP-hard~\cite{yu2013structure_nphard}.
On one side of this spectrum, decentralized methods such as~\cite{van2008reciprocal,ma2019searching}, plan paths for agents in their individual search spaces and can be leveraged to solve similar problems to MAPF-AA. These approaches scale well but can hardly guarantee completeness and optimality.
On the other side of the spectrum, centralized methods~\cite{standley2010finding} plan in the joint configuration space of agents, which guarantees optimality but scales poorly.
In the middle of the spectrum, methods like M*~\cite{wagner2015subdimensional}, Conflict-based Search (CBS) \cite{sharon2015conflict}, etc, begin by planning each agent an individual optimal path in a decoupled manner and couples agents for planning only when needed to resolve collisions. These methods guarantee optimality while bound the search space and thus scale relatively well.
This work limits its focus to planners with solution optimality guarantees.

In recent years, many variants of MAPF have been proposed, which span another spectrum from conventional MAPF~\cite{stern2019multi} to multi-agent motion planning (MAMP)~\cite{cohen2019optimal,shome2020drrt}, a generalized version of MAPF where the motion of agents are planned in continuous space and time. While being general to many applications, MAMP can be computationally expensive due to motion constraints, high degree-of-freedom of each agent, etc.
Within the spectrum, many different variants of MAPF have been proposed, each focus on relaxing different aspects of MAPF, such as shape of agents~\cite{li2019multi}, different moving speeds~\cite{walker2018extended,andreychuk2019multi}, multiple objectives~\cite{ren2021multi}, motion delays~\cite{ma2017multi}, etc.
A similar problem of MAPF-AA has been considered in~\cite{walker2018extended,andreychuk2019multi}.
In this work, we choose continuous-time CBS (CCBS)~\cite{andreychuk2019multi} as a baseline for our experiments. 

Among planners that solve conventional MAPF to optimality, there is no single planner that outperforms all others in all settings~\cite{felner2017search}.
To fuse the benefits of different MAPF planners, Meta-agent CBS (MA-CBS)~\cite{sharon2012meta} combines CBS with A*-based methods and has been shown to improve the performance.
However, due to the lack of any A*-based planner for MAPF-AA, we are not aware of any extension of MA-CBS for MAPF-AA that combines the benefits of both A*-based and CBS-based methods.
This work also fills this gap and our numerical results show that such fusion enhances the success rates of CCBS, the state-of-the-art, up to 12\%.

\section{Problem Description}\label{sec:problem}

Let index set $I = \{1,2,\dots,N\}$ denote a set of $N$ agents. All agents move in a workspace represented as a finite graph $G=(V,E)$ where the vertex set $V$ represents the possible locations of agents and the edge set $E =V \times V$ denotes the set of all possible actions that can move an agent $i$ between any two adjacent vertices in $V$.
An edge between $u,v\in V$ is denoted as $(u, v)\in E$.
In this work, we use a superscript $i \in I$ over a variable to represent the agent to which the variable belongs (e.g. $v^i\in V$ means a vertex corresponding to agent $i$).
Let $v_o^i, v_f^i\in V$ denote the start and goal vertices of agent $i$ respectively.

All agents share a global clock and start their motion at $v_o^i$ from time $t=0$. For each edge $e \in E$, let $D^i(e) \in \mathbb{R}^{+}$ denote the \emph{duration} for agent $i$ to go through edge $e$. Note that, for the same edge $e \in E$, durations $D^i(e),D^j(e)$ for two different agents $i,j \in I$ can be different. When agent $i$ goes through $(v_1,v_2) \in E$ between times $(t_1,t_1+D^i(v_1,v_2))$, agent $i$ occupies:
(1) vertex $v_1$ at time $t=t_1$, (2) vertex $v_2$ at time $t=t_1+D^i(v_1,v_2)$ and (3) both $v_1$ and $v_2$ for any time point within the open interval $(t_1,t_1+D^i(v_1,v_2))$.\footnote{We do not consider the case where edges criss-cross each other since this case can be handled by adding an additional vertex at the location where two edges criss-cross.}
Any two agents $i,j \in I$ are in conflict if they both occupy a same vertex at any time.

Let $\pi^i(v^i_{1}, v^i_{\ell})$ denote a path that connects vertices $v^i_{1}$ and $v^i_{\ell}$ via a sequence of vertices $(v^i_{1},v^i_{2},\dots,v^i_{\ell})$ in $G$, where any two vertices $v^i_{k}$ and $v^i_{k+1}$ are connected by an edge $(v^i_{k}, v^i_{k+1}) \in E$.
Let $g(\pi^i(v^i_{1}, v^i_{\ell}))$ denote the cost value associated with the path, which is defined as the sum of duration of edges along the path, $i.e.$ $g(\pi^i(v^i_{1}, v^i_{\ell})) = \Sigma_{k=1,2,\dots,{\ell-1}} D^i(v^i_{k},v^i_{k+1})$.
Without loss of generality, to simplify the notations, we also refer to a path $\pi^i(v^i_{o}, v^i_{f})$ for agent $i$ between its start and goal as simply $\pi^i$. 
Let $\pi=(\pi^1,\pi^2,\dots, \pi^N)$ represent a joint path for all the agents. Its cost is defined as the sum of the individual path costs over all the agents, $i.e.$, $g(\pi) = \Sigma_i g^i(\pi^i)$.

The objective of the multi-agent path finding with asynchronous actions (MAPF-AA) is to find a conflict-free joint path $\pi$ connecting $v^i_o,v^i_f$ for all agents $i \in I$ such that $g(\pi)$ is minimum.


\section{Loosely Synchronized Search}\label{sec:method}

\graphicspath{{contents/figures/}}


\vspace{-0.5mm}
\subsection{Notation and State Definition}\label{sec:method:pre}

Let $\mathcal{G}=(\mathcal{V},\mathcal{E}) = \underbrace{G \times G \times \dots \times G}_{\text{$N$ times}}$ denote the joint graph which is the Cartesian product of $N$ copies of $G$, where each $v \in \mathcal{V}$ represents a joint vertex and $e \in \mathcal{E}$ represents a joint edge that connects a pair of joint vertices.
The joint vertex corresponding to the starts and goals of agents is $v_o = (v^1_o,v^2_o,\cdots,v^N_o)$ and $v_f = (v^1_f,v^2_f,\cdots,v^N_f)$ respectively. 

In this work, a search state $s=(s^1,s^2,\dots,s^N)$ is defined to be a set of \emph{individual states} $s^i, \forall i\in I$ where each $s^i$ is a tuple of four components
\begin{itemize}
	\item $v(s^i) \in V$, an (individual) vertex in $G$;
	\item $p(s^i) \in V$, the parent vertex of $v(s^i)$, from which $v(s^i)$ is reached;
	\item $t(s^i)$, the timestamp of $v(s^i)$, representing the arrival time at $v(s^i)$ from $p(s^i)$;
	\item $t(p(s^i))$, the individual timestamp of $p(s^i)$, representing the departure time from $p(s^i)$ to $v(s^i)$.
\end{itemize}
Intuitively, $s^i = \{v(s^i), p(s^i), t(s^i), t(p(s^i))\}$ describes the location of agent $i$ within time interval $[t(p(s^i)), t(s^i)]$ with a pair of vertices $(p(s^i), v(s^i))$.
For the initial state, we define $p(s^i_o)=v(s^i_o)=v^i_o$ and $t(s^i_o)=t(p(s^i_o))=0, \forall i \in I$. 
Given two individual states $s^i_1,s^i_2$ of agent $i$, we say $s^i_1=s^i_2$ if each of the four elements in $s^i_1$ is equal to the counterpart in $s^i_2$.
For two states $s_1$ and $s_2$, $s_1=s_2$ if and only if $s^i_1=s^i_2, \forall i \in I$; otherwise, $s_1$ and $s_2$ are different states.

Following the definition of a conflict in Sec.~\ref{sec:problem}, given a state $s$, let $\Psi(s) \subseteq I$ represent a conflict checking function that checks the state $s$ for all pairs of agents $i,j \in I$ and returns a set of agents that are in conflict.
$\Psi(s)$ returns an empty set if no agent is in conflict in state $s$.

\vspace{-0.5mm}
\subsection{Algorithm Overview}\label{sec:method:algo_overview}
As in the well-known A* algorithm \cite{astar}, every state $s$ identifies a partial solution (path) $\pi(v_o,v(s))$ from $v_o$ to $v(s)$ and let $g(s)$ represent the cost of that partial solution. At any time of the search, let OPEN denote the priority queue containing candidate states, which are prioritized by their $f$-values $f(s) := h(s) + g(s)$, where $h(s)$ is the heuristic value that underestimates the cost-to-goal at $s$.

\begin{algorithm}[htbp]
	\caption{Pseudocode for A*, {\color{blue}LS-A*} }\label{alg:lss}
	\begin{algorithmic}[1]
		\State{add initial state $s_o$ to OPEN}
		\While{OPEN not empty} \Comment{Main search loop}
		
		\State{$s_k \gets$ OPEN.pop() }
		\State{\textbf{if} $v(s_k)=v_f$ \textbf{then}}
		\State{\indent \textbf{return} \text{Reconstruct($s_k$)}}
		\State{\color{blue}$S_{ngh} \gets$ \text{\it GetNgh}($s_k$) } 
		\State{{\color{blue}// LS-A* differs from A* in {\it GetNgh}($s_k$)}}
		\ForAll{$s_l \in S_{ngh}$ }
		\State{\textbf{if} $\Psi(s_l) \neq \emptyset$}
		\State{\indent \textbf{continue}}
		\State{\textbf{if} {\color{blue} Compare($s_l$)} \textbf{then}} \Comment{false = discard $s_l$}
		\State{{\color{blue}// LS-A* differs from A* in Compare($s_l$)}}
		\State{\indent $f(s_l) \gets g(s_l)$ + $h(s_l)$}
		\State{\indent add $s_l$ to OPEN}
		\State{\indent parent($s_l$) $\gets s_k$}
		\EndFor
		\EndWhile \label{}
		\State{\textbf{return} Failure}
	\end{algorithmic}
\end{algorithm}
\vspace{-1mm}

As shown in Algorithm \ref{alg:lss}, Loosely Synchronized A* (LS-A*) begins by adding initial state $s_o$ to OPEN. In each iteration (from line 2), the state $s_k$ with the minimum $f$-value is popped from OPEN. Then $v(s_k)$ is compared with $v_f$ and if $s_k$ \emph{visits} $v_f$ ($i.e.$ $v(s_k)=v_f$), then a conflict-free solution is identified and reconstructed by iteratively tracking the parent of states from $s_k$ to $s_o$. Otherwise, neighbors are generated from $s_k$ (line 6) by {\it GetNgh}($s_k$), a procedure that generates a set of neighboring states (successors) of $s_k$ (Sec.~\ref{sec:method:ngh_gen}).
For each generated neighbor $s_l$, if $s_l$ leads to conflicts (line 8), $s_l$ is discarded.
Otherwise, $s_l$ is verified in Compare($s_k$) (Sec.~\ref{sec:method:compare}), to decide whether $s_l$ should be kept.
If $s_l$ is kept, then the corresponding $f,g,h$ values of $s_l$ are updated and $s_l$ is inserted into OPEN.
When OPEN depletes, the algorithm reports failure and there is no solution for the problem.

\subsection{Neighbor Generation}\label{sec:method:ngh_gen}


The first key difference between LS-A* and A* is that, instead of letting all agents plan their next actions in each planning step as in A*, LS-A* uses timestamps of agents in a state to decide which agent(s) should plan the next action. The entire procedure can be described in four steps.

\noindent\underline{Step (1)} The minimum timestamp $t_{\min}(s_k)$ and the second minimum timestamps among all agents within $s_k$ are computed:
\begin{eqnarray}
\vspace{-1mm}
	t_{\min}(s_k) =& \min_{i\in I} t(s^i_k).\label{eqn:tmin} \\
	t_{\min2}(s_k) =& \min \{t(s^i_k) \;|\; t(s^i_k) \neq t_{\min}(s_k), i\in I \}. \label{eqn:tmin2}
\vspace{-1mm}
\end{eqnarray}
Note that, for any state $s_k$, $t_{\min}(s_k)$ always exists but $t_{\min2}(s_k)$ may not exist (if all timestamps are the same).

\noindent\underline{Step (2)} The subset of agent(s) $I_{t_{min}}(s_k) \subseteq I$ with timestamp(s) equal to $t_{min}(s_k)$ is computed:
\begin{equation}
\vspace{-1mm}
I_{t_{\min}}(s_k) = \arg\min_{i\in I} t(s^i_k).
\vspace{-1mm} \label{eqn:i_t_min}
\end{equation}
$I_{t_{\min}}(s_k)$ describes the subset of agents in $s_k$ that is allowed to plan their next actions.

\noindent\underline{Step (3)} We call an individual state $s^i_l$ generated from $s^i_k$ an \emph{individual neighbor} of $s^i_k$ and let $S^i_{ngh}(s^i_k)$ represent a set of individual neighbors of $s^i_k$. 
In this step, $S^i_{ngh}(s^i_k)$ is computed for each agent $i \in I$, given state $s_k$: 
\begin{itemize}
	\item For $i \notin I_{t_{min}}(s_k)$, agent $i$ is not allowed to plan actions and $S^i_{ngh}(s^i_k)$ contains only a copy of $s^i_k$.
	\item For $i \in I_{t_{min}}(s_k)$, agent $i$ plans actions, including both wait action and move actions, and $S^i_{ngh}(s^i_k)$ contains totally $(|Adj(v({s^i_k}))|+1)$ individual neighbors, where $Adj(u), u\in V$ represents the set of adjacent (individual) vertices of $u$ in graph $G$. 
\end{itemize}
Specifically, for action that moves agent $i$, for each vertex $u \in Adj(v^i_k)$ , a corresponding individual state $s^i_l = \{v(s^i_l), p(s^i_l), t(s^i_l), t(p(s^i_l)) \}$ is generated by
\vspace{-1mm}
\begin{eqnarray}
v(s^i_l) &=& u \label{eqn:v1}\\
p(s^i_l) &=& v(s^i_k) \label{eqn:p1}\\
t(s^i_l) &=& t(s^i_k) + D^i(v(s^i_k), v(s^i_l)) \label{eqn:duration1}\\
t(p(s^i_l)) &=& t(s^i_k) \label{eqn:tp1}
\vspace{-1mm}
\end{eqnarray}
where $D^i(v(s^i_k), v(s^i_l))$ denote the duration for agent $i$ to move through edge $(v(s^i_k), v(s^i_l))$. Then, the generated $s^i_l$ is added to $S^i_{ngh}(s^i_k)$. 
For action that makes agent $i$ wait, an individual state $s^i_l$ is generated by
\vspace{-1mm}
\begin{eqnarray}
v(s^i_l) &=& v(s^i_k)\\
t(s^i_l) &=& t(s^i_k) + D^i_{wait} \label{eqn:duration2} 
\vspace{-1mm}
\end{eqnarray} 
while $p(s^i_l)$ and $t(p(s^i_l))$ are generated by Equation (\ref{eqn:p1}) and (\ref{eqn:tp1}) respectively.
Here $D^i_{wait}$ denotes the amount of wait time and is computed as:
\begin{itemize}
	\item If $t_{\min2}(s_k)$ exists 
    \vspace{-1mm}
    \begin{eqnarray} D^i_{wait} = t_{\min2}(s_k)-t_{\min}(s_k), \label{eqn:wait_till_next}
    \vspace{-1mm}
    \end{eqnarray}
	\item Otherwise (all agents in $s_k$ have the same timestamps and $t_{\min2}(s_k)$ does not exist), 
    \vspace{-1mm}
	\begin{eqnarray} D^i_{wait} =  \min_{i\in I, e\in E} D^i(e). 
    \vspace{-1mm}
	\end{eqnarray}
\end{itemize}

\noindent\underline{Step (4)} $S_{ngh}$ is computed by taking combination of $S^i_{ngh}$ over all agents $i \in I$:
\vspace{-1mm}
\begin{equation}
	S_{ngh} = \{(s^1_l,s^2_l, \dots, s^N_l) \;|\; s^i_l \in S^i_{ngh}, \forall i \in I \}.
\vspace{-1mm}
\end{equation}

\noindent\underline{Remarks}
In {\it GetNgh}, wait action plays a key role in ``synchronizing'' subset of agents with Equation (\ref{eqn:wait_till_next}).
The wait action guarantees that, for each joint vertex $u \in \mathcal{V}$, after rounds of neighbor generation, there exists a state $s$ with $v(s)=u$ and $I_{t_{min}}(s) = I$ (Lemma~\ref{lem:sync_exist} in Sec.~\ref{sec:analysis}).
In such a state $s$, all timestamps of agents are the same and the algorithm needs to consider the actions of all agents together.
We term such a state a \emph{synchronized} state:
\begin{definition}
	A state $s$ is a synchronized state, if $t(s^i)=t(s^j), \forall i,j \in I, i\neq j$.
\end{definition}
In LSS, a state is either synchronized or asynchronized.
As we will see in Sec. \ref{sec:analysis}, the existence of synchronized states guarantees the completeness (not the optimality) of LS-A*.


\subsection{State Comparison}\label{sec:method:compare}
Different from A*, where a scalar $g$-value is used to compare states, in LS-A*, comparing states based solely on their $g$-values may not be enough: timestamps of agents in a state $s_k$ are relevant to potential conflicts along future paths from $s_k$. Thus, the timestamps of all agents in a state, which can be formulated as a vector, need to be properly handled for state comparison. This leads to the usage of \emph{dominance} \cite{ehrgott2005multicriteria} that compares two vectors.
\begin{definition}[Strict Dominance]
	For any two states $s_1$ and $s_2$ with the same joint vertex ($i.e.$ $v(s_1)=v(s_2)$), $s_1$ strictly dominates $s_2$, (notationally $ s_k \succ s_l $), if $t(s^i_1) < t(s^i_2), \forall i \in I$.
\end{definition}

In Sec. \ref{sec:combine_m}, we discuss the usage of other types of dominance.
For now, with strict dominance in hand, we introduce the Compare procedure, as shown in Algorithm \ref{alg:compare}. At each joint vertex $v \in \mathcal{V}$, a set of non-dominated states $\alpha(v)$ at $v$ is maintained. Initially, $\alpha(v) = \emptyset, \forall v \in \mathcal{V}\backslash \{v_o\}$ and $\alpha(v_o)$ contains only the initial state $s_o$. During the search, when a state $s_l$ is generated, to decide if $s_l$ should be pruned or not, $s_l$ is compared with every states in $\alpha(v(s_l))$. If $s_l$ is strictly dominated, $s_l$ is discarded. Otherwise, $s_l$ is added to $\alpha(v(s_l))$ and added to OPEN.

\begin{algorithm}
	\caption{Pseudocode for compare($s_l$)}\label{alg:compare}
	\begin{algorithmic}[1]
		\ForAll{$s_k \in \alpha(v(s_l))$}
		\State{\textbf{if} $ s_k \succ s_l$ \textbf{or} $s_k = s_l$ \textbf{then}}
		\State{\indent\Return false \Comment should be discarded}
		\EndFor
		\State{add $s_l$ to $\alpha(v(s_l))$}
		\State{\Return true \Comment should be added to open list}
	\end{algorithmic}
\end{algorithm}

\section{Analysis}\label{sec:analysis}

\ifthenelse{\boolean{shortver}}{%
    In this section, we list the key lemmas and show LS-A* is complete and optimal: LS-A* either computes an optimal solution or reports failure if no one exists. Detailed proof can be found in the full version of this work~\cite{ren2021loosely}.
}{
    In this section, we show LS-A* is complete and optimal: LS-A* either computes an optimal solution or reports failure if no one exists.
}

\begin{corollary}\label{lem:t_min_increase}
	Let $s_l$ represent a neighbor state generated from $s_k$, then $t_{\min}(s_l) > t_{\min}(s_k)$.
\end{corollary}
This corollary comes from the construction of {\it GetNgh}($s_k$) (In Eqn.~\ref{eqn:duration1}, \ref{eqn:duration2}, durations are always strict positive). 

\begin{lemma}\label{lem:sync_exist}
	For a state $s$, there exists a descendent state $s_l$ from $s$ with $v(s_l) = v(s)$ and $s_l$ is a synchronized state.
\end{lemma}

\ifthenelse{\boolean{shortver}}{%
}{
\begin{proof}
	In {\it GetNgh}($s$), for every agent $i \in I_{t_{\min}}(s)$, the wait action makes agent $i$ stay at the same vertex while increase the timestamp to $t_{\min2}(s)$ (If $t_{\min2}(s)$ does not exists, then $s$ is itself a synchronized state and the Lemma holds). Let $t_{\max}(s)=\max_{i \in I} (t(s^i))$ represent the maximum timestamps over agents in state $s$. Now, consider the neighbor state $s_k \in S_{ngh}$ generated from $s$ by letting all agents $i \in I_{t_{\min}}(s)$ choose the wait action, then $v(s_k) = v(s)$ and $t_{\min}(s) < t_{\min}(s_k) \leq t_{\max}(s) = t_{\max}(s_k)$. In addition, state $s_k$ is not strictly dominated by $s$ and is thus not pruned. Repeating the above process results in a descendant state $s_l$ with $v(s_l)=v(s)$ and $t_{\min}(s_l) = t_{\max}(s) = t_{\max}(s_l)$. As $t_{\min}(s_l) \leq t(s^i_l) = t(s^j_l) \leq t_{\max}(s_l), \forall i,j \in I$, $s_l$ is a synchronized state.
\end{proof}
}

\begin{corollary}\label{lem:expand}
	If state $s$ is a synchronized state, then {\it GetNgh}($s$) \emph{expands}\footnote{A node in a graph is expanded if all of its neighbor nodes are generated (visited). See \cite{pearl1984intelligent} for more details.} joint vertex $v(s)$ in $\mathcal{G}$: let $S_{ngh}(s)$ denote the set of neighbor states returned by {\it GetNgh}($s$), for every adjacent joint vertex $u$ of $v(s)$ in $\mathcal{G}$, there exists a neighbor state $s_l \in S_{ngh}(s)$ such that $v(s_l) = u$.
\end{corollary}

\begin{lemma}\label{lem:finite_num_states}
	For each joint vertex $v_k \in \mathcal{V}$, there exists only a finite number of states $s$ with $v(s) = v_k$.
\end{lemma}

\ifthenelse{\boolean{shortver}}{%
}{
\begin{proof}
	Joint graph $\mathcal{G}$ is finite and there exists only a finite number of partial solutions from start to a joint vertex $v_k \in \mathcal{V}$ unless agent wait infinitely at $v_k$.
	From lemma \ref{lem:sync_exist}, for every joint vertex $v_k \in \mathcal{V}$, there exists a corresponding synchronized state $s_k$ generated by LS-A* with $v(s_k)=v_k$. From lemma \ref{lem:t_min_increase} and strict dominance pruning rules in Algorithm \ref{alg:compare}, any descendant state $s_k'$ from $s_k$ with $v(s_k')=v(s_k)$ are pruned. In addition, any states $s_l$ with $v(s_l)=v(s_k)$ and $t_{\min}(s_l) > t_{\min}(s_k)$ ared pruned. Therefore, agents cannot wait infinitely at a joint vertex.
\end{proof}
}

\begin{theorem}\label{thm:complete}
	LS-A* is complete.
\end{theorem}

\ifthenelse{\boolean{shortver}}{%
}{
\begin{proof}
	From Lemma \ref{lem:finite_num_states}, if there is no solution, LS-A* terminates in finite time when OPEN depletes and report failure. 
	If there is a solution, from Lemma \ref{lem:sync_exist} and Corollary \ref{lem:expand}, every joint vertex in graph $\mathcal{G}$ is expanded until LS-A* finds a solution. 
\end{proof}
}

\begin{corollary}\label{lem:dom}
	For two states $s_k$ and $s_l$ with $v(s_k)=v(s_l)$, if $s_k$ strictly dominates $s_l$, $s_l$ can not leads to a solution with smaller cost than $s_k$.
\end{corollary}

\begin{corollary}\label{lem:no_change_in_middle}
	Given a state $s_k$ and a neighbor state $s_l$ generated from {\it GetNgh}($s_k$), agent $i \in I$ occupies both $v(s^i_k)$ and $p(s^i_k)$ for any time between $t_{min}(s_k)$ and $t_{min}(s_l)$. 
\end{corollary}

\ifthenelse{\boolean{shortver}}{%
}{
This corollary follows from the conflict definition in the problem description and the state definition in Sec.~{\ref{sec:method}}. Given $s^i=\{v^i, p(s^i), t(s^i), t(p(s^i))\}$, the vertices occupied by agent $i$ does not change at any time between $(t(p(s^i)),t(s^i))$. In other word, the vertices occupied by agents changes only at timestamps $t(p(s^i))$ and $t(s^i)$. For state $s_k$ and its neighbor $s_l \in S_{ngh}(s_k)$, there is no timestamps between $( t_{min}(s_k), t_{min}(s_l) )$ and therefore there is no change in vertices occupied by agents.
}

\begin{lemma}\label{lem:leave_in_middle}
	When generating neighbors for a state $s$, for any agent $i \in I_{t_{min}}(s)$, waiting for an amount of time in $(0, D^i_{wait})$ does not leads to any solution with smaller cost.
\end{lemma}

\ifthenelse{\boolean{shortver}}{%
}{
\begin{proof}
	Let $s'$ be a state that is generated from $s$ by letting an agent $i\in I_{t_{min}}(s)$ wait for an amount of time in $(0, D^i_{wait})$. Based on Corollary \ref{lem:no_change_in_middle}, if a joint vertex cannot be reached from $s$ because of agent-agent conflicts, then this joint vertex cannot be reached from $s'$ as well. Therefore, for every state $s_l'$ generated from $s'$, there must be a corresponding state $s_l$ generated from $s$ with $v(s_l)=v(s_l')$ and $g(s_l) < g(s_l')$. Thus, for any solution that goes through $s'$, there exists a corresponding solution via $s$ with a smaller $g$-value.
\end{proof}
}

\begin{theorem}\label{thm:optimal}
	If there are solutions, LS-A* finds the one with the minimum $g$-value.
\end{theorem}

\ifthenelse{\boolean{shortver}}{%
}{
\begin{proof}
	From Corollary \ref{lem:leave_in_middle}, {\it GetNgh} procedure generates all possible neighbors of a state that can be part of an optimal solution.
	From Lemma \ref{lem:dom}, frontier set $\alpha(v)$ keeps track of all possible states at joint vertex $v$ that can be part of an optimal solution. All states in $\alpha(v)$ for any $v\in \mathcal{V}$ are inserted into OPEN. LS-A* selects candidate state from OPEN with the minimum $g$-value (same as A*) and therefore identifies the solution with the minimum cost.
\end{proof}
}

\section{Discussion and Extensions}\label{sec:combine_m}

\subsection{Switch Between Dominance Rules}
The aforementioned {\it GetNgh} procedure and strict dominance guarantee the existence of a synchronized state at each joint vertex.
After a synchronized state $s$ at $v(s)$ is generated and added into OPEN, for any descendant states $s_l$ with $v(s_l) = v(s)$, however, the algorithm can switch to a pruning rule with relaxed conditions instead on relying on strict dominance. This is helpful since more states, that are not part of an optimal solution, can be pruned. The relaxed conditions are defined through weak dominance~\cite{przybylski2017multi} as follows:
\begin{definition}[Weak Dominance]
	For any two states $s_1$ and $s_2$ with the same joint vertex ($i.e.$ $v(s_1)=v(s_2)$), $s_1$ weakly dominates $s_2$, if $t(s^i_1) \leq t(s^i_2), \forall i \in I$.
\end{definition}
With both the dominance rules, the algorithm can switch between them to decide whether a state $s_k$ should be pruned or not. If a synchronized state $s_l$ with $v(s_l)=v(s_k)$ has already been generated and inserted into OPEN during the search, then $s_k$ is compared with every state in $\alpha(v(s_k))$ with weak dominance.
Otherwise (which means no synchronized state has been generated at $v(s_k)$), $s_k$ is compared with every state in $\alpha(v(s_k))$ with strict dominance. Switching between the two dominance rules do not affect the proof, and thus the properties of LS-A* still hold.

\ifthenelse{\boolean{shortver}}{%
    
\subsection{Combination with other algorithms}
To demonstrate the generality of the proposed LSS, we combine LSS with M* and rM*~\cite{wagner2015subdimensional} (two A*-based algorithms for conventional MAPF) and propose LS-M* and LS-rM* for MAPF-AA.
In addition, we also extend MA-CBS~\cite{sharon2012meta,boyarski2015icbs} to combine the advantages of both CBS-based and A*-based planners.
More details can be found in~\cite{ren2021loosely}.

}{
    
\subsection{Relationship to A*}

With the problem definition in Sec.~\ref{sec:problem}, if all actions for all agents take the same amount of time, $i.e.$ $D^i(e_k)=D^j(e_l), \forall i,j \in I, \forall e_k,e_l \in E$, the MAPF-AA problem is then equivalent to a conventional MAPF problem. 
In this case, LS-A* is equivalent to regular A* for the following two reasons.
\begin{itemize}
	\item The timestamps of all agents in every state are the same (and every state is thus a synchronized state). As a result, in {\it GetNgh} procedure, all agents always plan their actions together.
	\item Since every state is synchronized, all agents share the same timestamp in each state, which can be described as a scalar; Besides, the algorithm always uses weak dominance, which is equavalent to ``$\leq$'' (no larger than) relationship between two scalar values as in regular A*.
\end{itemize}
Those two statements also explain how A* is extended to LS-A* with the two aforementioned procedures to handle asynchronous actions. 

\subsection{Relationship to Operator Decomposition}
When applying A* to conventional MAPF (without asynchronous actions), the number of neighbors generated in each iteration grows exponentially with respect to the number of agents.
Operator decomposition (OD)~\cite{standley2010finding} mitigates this challenge by generating \emph{intermediate} states, where an order between agents is established and only one agent is allowed to plan its next actions for each iteration. 
This order is, in general, established by the $f$-value of states.
When all agents have chosen their actions following the order, a \emph{standard} state is generated.
Both intermediate and standard states are treated in the same best-first search manner as in A*: both types of states are inserted into OPEN and selected based on $f$-values for expansion.
By doing so, OD avoids the generation of high cost states which may never be expanded.

From the perspective of OD, LSS is similar by allowing only a subset of agents to plan their next actions each time.
Additionally, the aforementioned synchronized states are equivalent to standard states in OD in a sense that all agents have planned their actions.
However, in LSS, the order between agents is established by timestamps of agents.

\subsection{Extension with M*}
To demonstrate the generality of the proposed LSS, we combine LSS with M*, an A*-based algorithm for conventional MAPF, and propose LS-M* for MAPF-AA.
Specifically, M* uses two concepts: the \emph{individual policy} and the \emph{collision set}. 
The individual policy of agent $i$ maps a vertex $v^i \in V$ to the next vertex along an optimal path connecting $v^i$ and $v^i_f$ ignoring any other agents.
With an individual policy, an agent is constrained to a one-dimensional search space from any vertex to its goal.
The collision set $I_C(v), v\in \mathcal{V}$ describes the subset of agents that are in conflict along paths through joint vertex $v$. 
Collision sets are initially all empty sets for all joint vertices and are enlarged during the search via (1) conflict detection, which detects collision between agents at joint vertices, and (2) back-propagation: when $I_C(v)$ of some joint vertex $v$ is enlarged, the collision sets of all joint vertices that are relevant to $v$ are also enlarged. 
To expand a joint vertex $v$, M* let agents $i \notin I_C(v)$ follow their individual policies and let agents $i \in I_C(v)$ consider all possible actions at $v^i$.
By doing so, M* plans in a ``compact'' search space with varying dimensionality embedded in $\mathcal{G}$.

To combine LSS with M* to handle MAPF-AA, similar procedures, as presented in LS-A*, is required.
First of all, we define the same search state as the one presented in the aforementioned LS-A* and a collision set $I_C(s)$ is defined for every state.
Secondly, when generating neighbors of a given state $s$, Steps (1), (2) and (4) , as stated in LS-A*, remain the same. Step (3) needs some adaption to consider collision set $I_C(s)$: for agents $i \in I_{t_{\min}}(s)$, if $i \notin I_C(s)$, agent $i$ is only allowed to follow its individual policy; otherwise, individual neighbors are generated in the same way as in LS-A*.
Finally, states are also compared and pruned by (strict and weak) dominance rules as in LS-A*, and when a state $s_k$ is dominated by $s_l$, $I_C(s_l)$ need to be back-propagated to $s_k$ so that the low dimensional search space embedded in $\mathcal{G}$ are properly maintained~\cite{ren21momstar}.

\subsection{Extension with Recursive M*}

As M* perform coupled planning for all agents in a collision set, recursive M* (rM*), a variant of M*, further extend the idea by (1) identifying spatially separated subsets of agents within a collision set and (2) performing coupled planning for each of these spatially separated subsets. During the search, rM* finds optimal paths for each of those subsets via a recursive call to rM* and, when expanding agents within each subset, agents are only allowed to follow those planned paths~\cite{wagner2015subdimensional}. Extending rM* to LS-rM* takes exactly the same procedures as extending M* to LS-M*.

\subsection{Extension with MA-CBS}

Among algorithms that solve conventional MAPF, meta-agent conflict-based search (MA-CBS)~\cite{sharon2012meta} and its improved version~\cite{boyarski2015icbs} combines the advantages of both CBS and A*-based planners. 
The search space of CBS grows exponentially with the number of conflicts detected~\cite{sharon2015conflict}. 
MA-CBS mitigates this burden by tracking the number of conflicts between any pair of agents and merging those agents as an meta-agent if the number of conflicts between them exceed some pre-defined threshold $B$. 
For each meta-agent, MA-CBS leverages A*-based algorithms to plan (joint) path.
MA-CBS thus fuses A*-based approach and CBS approach: when $B=0$, all agents are always merged as a single meta-agent and MA-CBS is equivalent to a pure A*-based approach; when $B=\infty$, agents are never merged and MA-CBS is equivalent to a pure CBS approach; when $0 < B < \infty$, MA-CBS combines both.
The test results~\cite{sharon2012meta} explores different threshold $B$ and show that MA-CBS outperforms CBS in many different scenarios.

In this work, we also consider extending MA-CBS for MAPF-AA by combining the proposed LSS approaches, such as LS-rM*, with the continuous-time CBS (CCBS) algorithm~\cite{andreychuk2019multi}, which extends CBS and handles MAPF-AA by using SIPP~\cite{phillips2011sipp} algorithm as the low level planner in CBS. 
With $B\in [0,\infty]$, MA-CBS algorithm varies between LS-rM* and CCBS.
Our experiments (Sec.~\ref{sec:result}) show that such combination improves CCBS under different scenarios.

}

\section{Numerical Results}\label{sec:result}

\graphicspath{{contents/figures/}}

\begin{table*}[htbp]
	\centering
	\tabcolsep=0.2cm
	\renewcommand{\arraystretch}{1.1}
	\vspace{2mm}
	\begin{tabular}{ | l | l | l | l | l | l | l | l | }
		\hline
		& & \multicolumn{6}{c}{Success Rates (Avg. Run Time in Seconds) } \vline
		\\ \hline
		Grids & B & N=2 & N=4 & N=8 & N=12 & N=16 & N=20 
		\\ \hline
		\multirow{4}{*}{\includegraphics[width=0.07\linewidth]{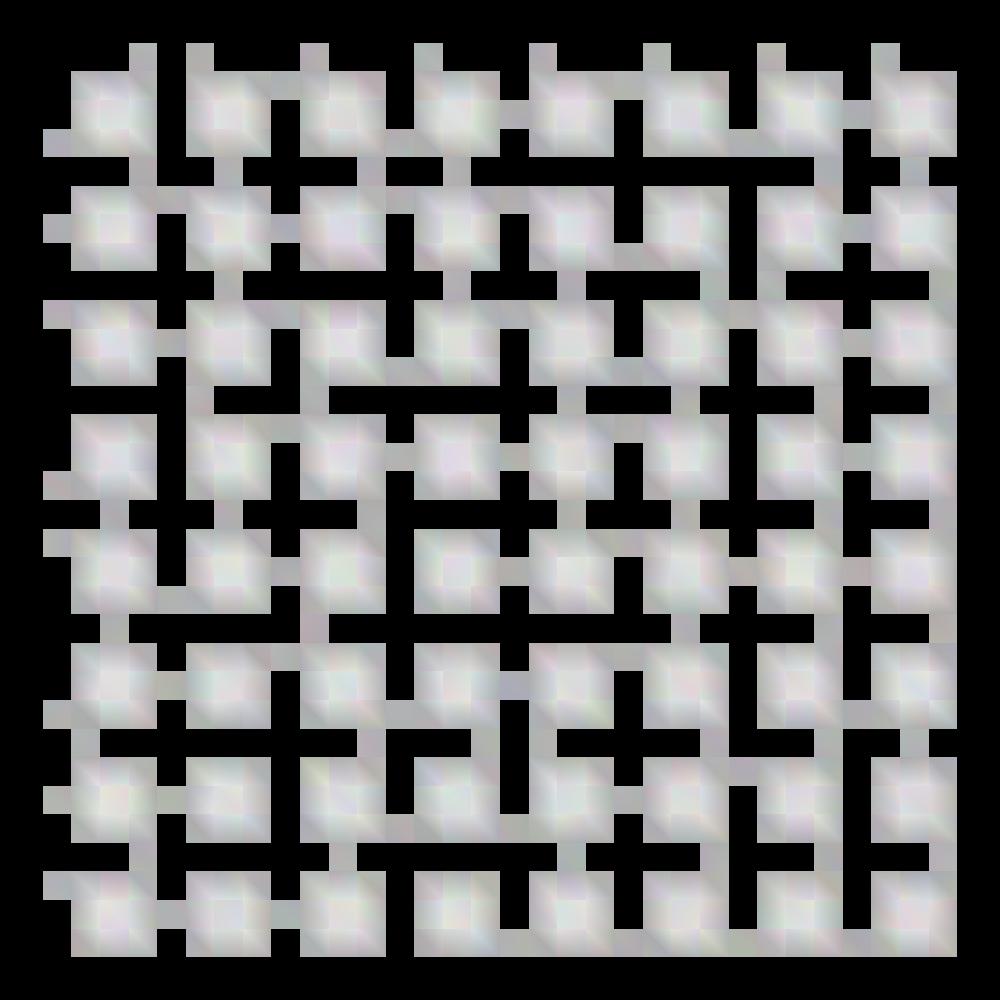}} 
		& 0 (LS-rM*) & 1.00 (0.17) & 0.84 (50.9) & 0.40 (183.3) & 0.04 (288.3) & 0 (-) & 0 (-) 
		\\ 
		& 1 & 1.00 (0.20) & 0.84 (53.0) & 0.44 (170.6) & 0.08 (276.11) & 0 (-) & 0 (-) 
		\\ 
		& 10 & 1.00 (0.019) & \textbf{0.92 (28.9)} & 0.60 (122.5) & 0.20 (241.1) & 0 (-) & 0 (-)  
		\\ 
		& 100 & 1.00 (0.019) & 0.92 (29.3) & \textbf{0.68 (99.0)} & \textbf{0.28 (227.3)}  & 0.04 (294.3) & 0 (-)
		\\ 
		(32x32) & $\infty$ (CCBS) & \textbf{1.00 (0.005)} & 0.88 (36.0) & 0.56 (138.5) & 0.24 (232.3)  & \textbf{0.04 (290.6)} & 0 (-) 
		\\ 
		\hline
		\multirow{4}{*}{\includegraphics[width=0.07\linewidth]{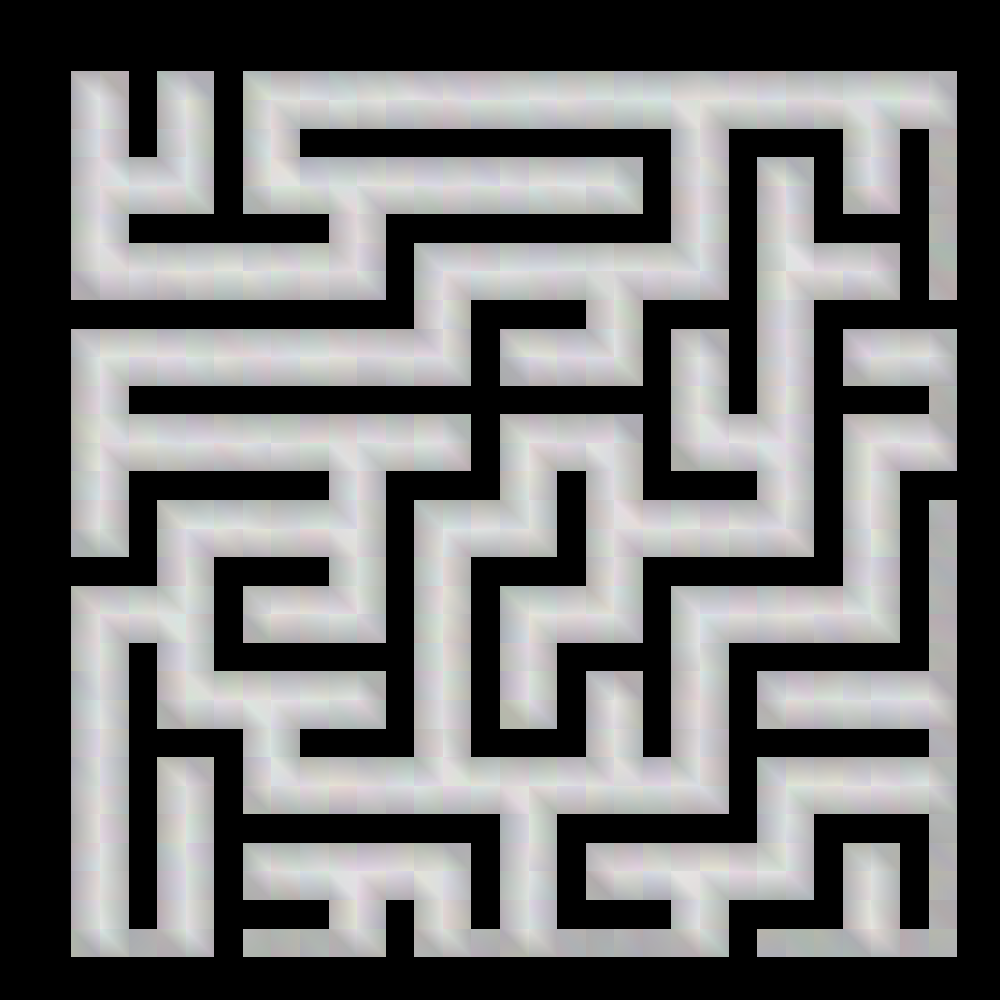}} 
		& 0 (LS-rM*) & 1.00 (0.11) & 0.84 (78.9) & 0.08 (278.8) & 0 (-) & 0 (-) & 0 (-) 
		\\ 
		& 1 & 1.00 (0.15) & 0.84 (57.9) & 0.08 (278.7) & 0 (-) & 0 (-) & 0 (-)  
		\\ 
		& 10 & 1.00 (0.08) & 0.88 (45.6) & 0.12 (265.8) & 0 (-) & 0 (-) & 0 (-) 
		\\ 
		& 100 & 1.00 (0.06) & 0.92 (25.2) & \textbf{0.32 (217.3)} & \textbf{0.04 (293.7)} & 0 (-) & 0 (-) 
		\\ 
		(32x32) & $\infty$ (CCBS) & \textbf{1.00 (0.03)} & \textbf{0.92 (24.5)} & 0.24 (241.9) & 0 (-) & 0 (-) & 0 (-) 
		\\
		\hline
		\multirow{4}{*}{\includegraphics[width=0.07\linewidth]{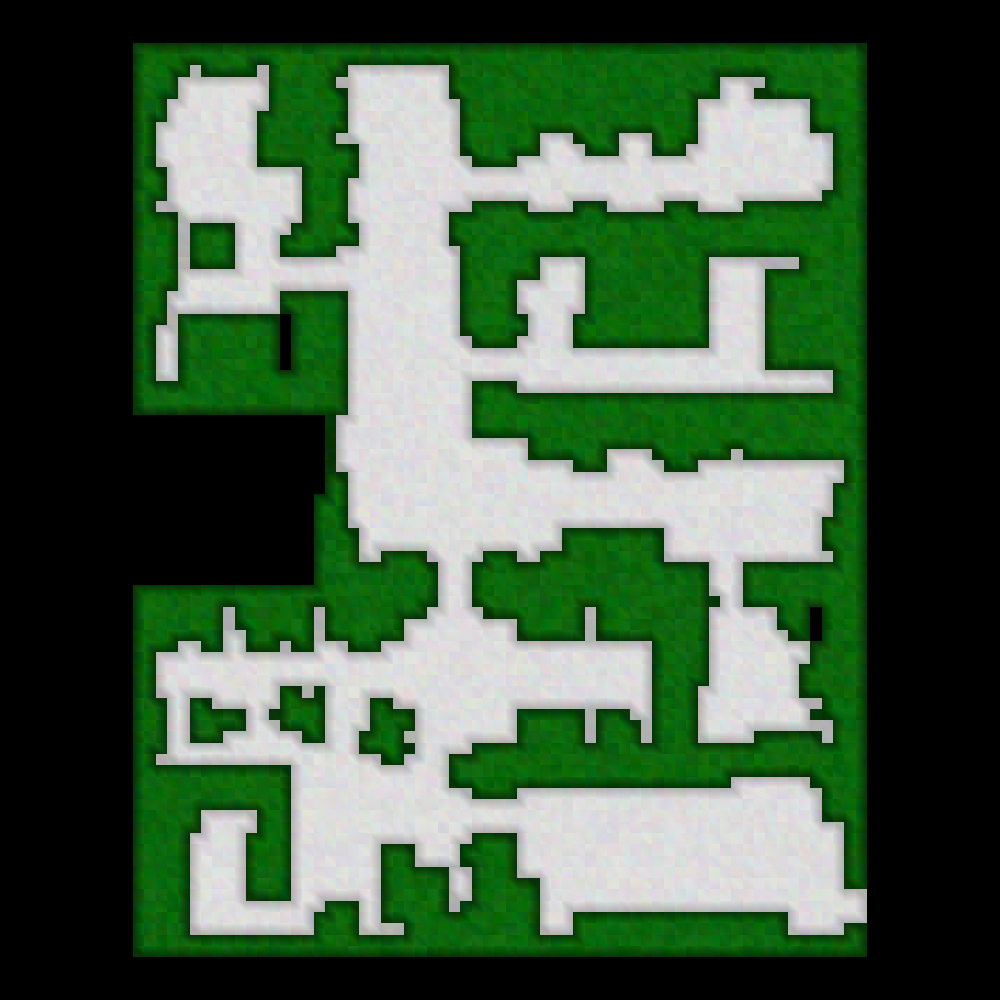}} 
		& 0 (LS-rM*) & 0.92 (14.7) & 0.80 (65.1) & 0.60 (137.1) & 0.04 (288.0) & 0 (-)  & 0 (-) 
		\\ 
		& 1 &\textbf{ 0.96 (19.9)} & 0.80 (60.4) & 0.60 (127.9) & 0.04 (297.1) & 0 (-)  & 0 (-)  
		\\ 
		& 10 & 0.96 (20.2) & 0.92 (24.3) & 0.68 (99.5) & 0.20 (245.0) & 0 (-) & 0 (-) 
		\\ 
		& 100 & 0.96 (20.1) & 0.92 (24.2) & \textbf{0.88 (45.4)} & \textbf{0.48} (185.1) & 0.16 (277.9) & 0.04 (295.9) 
		\\ 
		(65x81) & $\infty$ (CCBS) & 0.92 (24.0) & \textbf{0.92 (24.1)} & 0.84 (54.3) & 0.44 \textbf{(184.2)} & \textbf{0.16 (270.0)} & \textbf{0.04 (290.5)}
		\\
		\hline
	\end{tabular}
	\caption{Numerical results of MA-CBS with different merging threshold B.}
	\label{tab:macbs}
\end{table*}

\begin{figure*}[htbp]
	\centering
	\vspace{-5mm}
	\includegraphics[width=\linewidth]{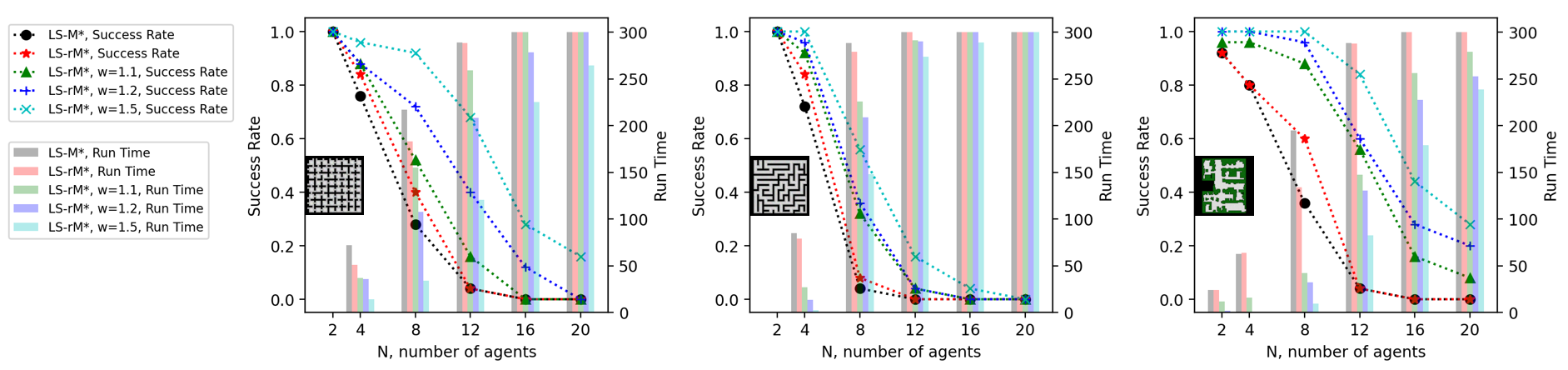}
	\vspace{-7mm}
	\caption{Average success rates of A*-based algorithms for finding optimal solution within one minute.}
	\vspace{-2mm}
	\label{fig:heu_weight}
\end{figure*}

All the algorithms were implemented in Python and tested on a computer with an Intel Core i7 CPU and 16 GB RAM.
We selected maps (grids) from~\cite{stern2019multi} and generated an un-directed graph by making each grid four-connected. 
The run time limit for each test instance is {\it five} minutes.
We report the performance of the proposed LSS approach with the following experiments.
First, we compare LS-A* and ``naive-A*'' (explained in Sec.~\ref{sec:lsa_vs_naivea}) with different durations to verify whether LS-A* saves computational effort when actions are asynchronous.
Second, we verify the performance of MA-CBS, using LS-rM* as the underlying meta-agent planner, by varying the merging threshold $B$.
Finally, we tested LS-rM* with varying heuristic inflation rates to learn how LS-rM* trades off between optimality and search efficiency.

\subsection{Naive A* and LS-A*}\label{sec:lsa_vs_naivea}

Naive-A* assumes the existence of a common time unit $\tau$ and a maximum possible time $T$, and discretizes the time dimension into a finite number of time steps $\{0,1,\dots,T/\tau\}$. This discretization guarantees that the actions of all agents begin/end concurrently.
Naive-A* conducts A* search in a time-augmented graph by visiting all possible time steps.
In our tests, the durations of edges were implemented as $D^i(e)=d^i, \forall e\in E$, where $d^i$ is a random integer sampled from $[1,K]$, with $K=10,100,1000$, representing the ``degree'' of asynchronous actions.
Note that, different agents $i,j$ can have $d^i \neq d^j$.
We fixed the number of agents with $N=2$ and ran tests in a $16\times16$ obstacle-free grid.

From Table~\ref{table:lsa}, LS-A* outperforms naive-A* in terms of number of states expanded as well as run time on average regardless of $K$. Additionally, when $K$ varies, LS-A* remains steady against those two metrics.
The results show the benefits of LS-A* as it avoids too fine a discretization of the time dimension.

\begin{table}[htbp]
	\centering
	\vspace{-0mm}
	\begin{tabular}{ |c|c|c|c| } 
		\hline
		\multicolumn{4}{|c|}{Avg. No. States Expanded (Avg. Run Time in Seconds)} \\
		\hline
		K & 10 & 100 & 1000 \\
		\hline
		Naive-A* & 542.9 (0.15) & 3148.3 (3.20) & 10839.0 (55.16) \\
		\hline
		LS-A* & {\bf 365.8 (0.09)} & {\bf 453.3 (0.08)} & {\bf 449.9 (0.08)}  \\ 
		\hline
	\end{tabular}
	\caption{Average number of states expanded and run time for Naive-A* and LS-A* with different duration functions. }
	\vspace{-5mm}
	\label{table:lsa}
\end{table}

\subsection{Meta-agent Conflict-based Search}

Table~\ref{tab:macbs} shows the results of the improved MA-CBS~\cite{boyarski2015icbs}, which uses SIPP and LS-rM* as low level planners, with a merging threshold $B \in \{0,1,10,100,\infty\}$.
When the number of ``internal'' conflicts between a pair of agents exceeds $B$, those two agents are merged as a meta-agent.
Note that when $B=0$, MA-CBS is the same as LS-rM* since all agents are always merged as one meta-agent, and when $B=\infty$, MA-CBS is the same as CCBS~\cite{andreychuk2019multi} since all agents are never merged.
Here, durations are set in the same way as in \ref{sec:lsa_vs_naivea} with $K=100$.
We select three grids from different categories (room, maze, game map) from~\cite{stern2019multi} and report the success rates of finding a solution within the time limit, as well as average run times (over all instances, both solved and unsolved) for a different number of agents $N\in \{2,4,8,12,16,20\}$.
The best performance for each $N$ is highlighted in \textbf{bold} text.

MA-CBS shows its benefits over CCBS and the maximum improvement is achieved in room and maze-like grids (the first and second maps) with $N=8, B=100$, where success rates are both enhanced by 12\% and the average run time is shortened.
It is also worthwhile to note that the selection of $B$ remains an open question, as different $B$ can affect the performance of the algorithm in various environments.

\subsection{Heuristic Inflation}

For A*-based algorithms, a well-known technique that trades off between bounded sub-optimality and search efficiency is using inflated heuristics~\cite{pearl1984intelligent}: $f=g+w\cdot h$, where $w \geq 1$ is the inflation rate.
In general, $w>1$ makes A* find a bounded sub-optimal solution faster.
As shown in Fig.~\ref{fig:heu_weight}, we plotted the success rates and run time of LS-rM* by varying the inflation rate $w \in \{1.1, 1.2, 1.5\}$. We also show the results of LS-M* and LS-rM* without inflation $i.e.$ $w=1.0$ as baselines. 
It is obvious that heuristic inflation helps in improving success rates and average run times in all grids tested.

\subsection{Real Robot Test}

We verify the proposed inflated LS-rM* in the Robotarium~\cite{wilson2020robotarium}, a remotely accessible swarm robotics research platform, by simulating and executing the planned paths, as shown in the video\footnote{\url{https://drive.google.com/file/d/1EX5CcOA6oUCmYX3BD3Zdi2iADoA9sMuH/view?usp=sharing} }.

\section{Conclusion}\label{sec:conclude}
We proposed an approach named Loosely Synchronized Search that can convert A*-based planners to a version that can solve the multi-agent path finding (MAPF) problem with asynchronous actions. 
We proved the theoretical properties of LSS and presented extensive numerical results to verify its performance against the state of the art MAPF algorithms. Possible future work includes applying LSS with other A*-based algorithms, such as EPEA* \cite{goldenberg2014enhanced}, or further extend LSS to other variants of MAPF.

\bibliographystyle{plain}
\bibliography{ref}


\end{document}